\documentclass{article}
\usepackage{ijcai24}

\usepackage{times}
\usepackage{soul}
\usepackage{url}
\usepackage[hidelinks]{hyperref}
\usepackage[utf8]{inputenc}
\usepackage[small]{caption}
\usepackage{graphicx}
\usepackage{amsmath}
\usepackage{amsthm}
\usepackage{booktabs}
\usepackage{algorithm}
\usepackage{algorithmic}
\usepackage[switch]{lineno}
\usepackage{color}
\usepackage{amssymb}
\usepackage{multirow}
\usepackage{bbm}

\title{EfficientGS: Streamlining Gaussian Splatting for Large-Scale High-Resolution Scene Representation}

\iftrue
\author{
Wenkai Liu$^1$,
Tao Guan$^1$,
Bin Zhu$^2$,
Lili Ju$^3$,
Zikai Song$^1$,
Dan Li$^1$,
Yuesong Wang$^1$\And
Wei Yang$^1$
\affiliations
$^1$Huazhong University of Science and Technology\\
$^2$Wuhan Farsee2 Technology Co., Ltd.\\
$^3$University of South Carolina
\emails
\{wenkai\_liu, qd\_gt, skyesong, lidan, yuesongwang,weiyangcs\}@hust.edu.cn,
binzhu@farsee2.com,\\
ju@math.sc.edu
}
\fi

\begin{document}

\maketitle
\begin{abstract}
In the domain of 3D scene representation, 3D Gaussian Splatting (3DGS) has emerged as a pivotal technology. However, its application to large-scale, high-resolution scenes (exceeding 4k$\times$4k pixels) is hindered by the excessive computational requirements for managing a large number of Gaussians. Addressing this, we introduce 'EfficientGS', an advanced approach that optimizes 3DGS for high-resolution, large-scale scenes.
We analyze the densification process in 3DGS and identify areas of Gaussian over-proliferation. We propose a selective strategy, limiting Gaussian increase to key primitives, thereby enhancing the representational efficiency. Additionally, we develop a pruning mechanism to remove redundant Gaussians, those that are merely auxiliary to adjacent ones.

For further enhancement, we integrate a sparse order increment for Spherical Harmonics (SH), designed to alleviate storage constraints and reduce training overhead. Our empirical evaluations, conducted on a range of datasets including extensive 4K+ aerial images, demonstrate that 'EfficientGS' not only expedites training and rendering times but also achieves this with a model size approximately tenfold smaller than conventional 3DGS while maintaining high rendering fidelity.
\end{abstract}

\begin{figure}
    \centering
    \includegraphics[width=1\linewidth]{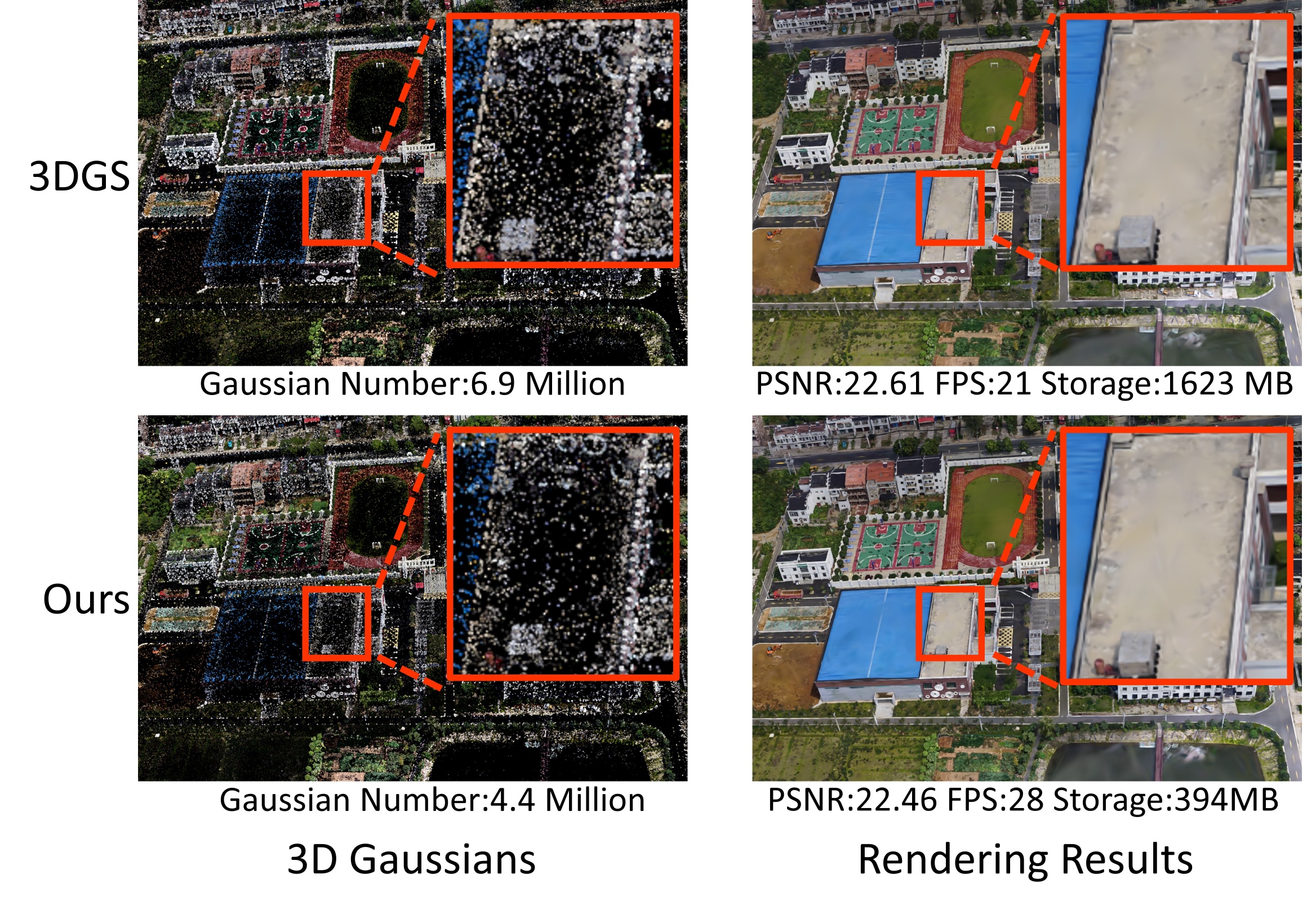}
    \caption{\textbf{Comparison with vanilla 3DGS and our EfficientGS on a large-scale high-resolution scene.} The vanilla 3DGS suffers from its redundancy and thus has a low FPS and high storage pressure when representing large-scale high-resolution scenes, whereas our approach allows for a more efficient scene representation in contrast.}
    \label{fig:head}
\end{figure}

\section{Introduction}

3D scene representation serves as a pivotal mechanism for the digital reconstruction and interpretation of the real world within computational environments. This field, particularly in the context of large-scale scene representation, has garnered significant attention due to its diverse and impactful applications. These applications span a broad spectrum, including autonomous driving~\cite{huang2022multi}, aerial surveying~\cite{bozcan2020air}, embodied artificial intelligence~\cite{duan2022survey}, digital mapping technologies~\cite{GoogleEarth}, and urban planning\cite{souza2022city}, among others.

Traditionally, large-scale scene representations have predominantly employed mesh-based methods. The conventional reconstruction pipeline for these methods typically comprises several stages: Structure-From-Motion (SFM)~\cite{schonberger2016structure}, Multi-View Stereo (MVS)~\cite{schonberger2016pixelwise,xu2021exploiting}, surface-based reconstruction~\cite{labatut2009robust}, and texture mapping~\cite{waechter2014let}. While mesh-based approaches offer the advantages of ease in editing and rendering, they exhibit inherent limitations in handling non-Lambertian surfaces. Such limitations often result in reconstructed surfaces plagued with artifacts, including holes, collapses, and floating discrepancies.

A significant advancement in this domain has been the introduction of the Neural Radiance Field (NeRF) by Mildenhall et al.~\cite{mildenhall2020nerf}. NeRF innovatively represents scenes as radiance fields and employs Multilayer Perceptrons (MLP) for implicit encoding. This methodology facilitates photo-realistic rendering through differentiable volume rendering, representing a notable leap in rendering quality. Capitalizing on NeRF's robust representation capabilities, subsequent studies~\cite{MegaNerf,tancik2022block,xu2023grid} have extended its application to modeling large-scale scenes.
However, the adoption of NeRF-based approaches is not without challenges. The implicit representation strategy inherent in NeRF introduces significant computational overheads, particularly in the contexts of training and rendering. This aspect poses a critical limitation, calling for further research to optimize the balance between rendering quality and computational efficiency.

The recent introduction of 3D Gaussian Splatting (3DGS), as explored in Kerbl et al.~\cite{kerbl20233d}, represents a significant advancement in scene representation. This method utilizes 3D Gaussians, initiated from sparse SFM points, to depict scenes, thereby streamlining training and rendering by avoiding computations in empty spaces. The integration of anisotropic covariance and Spherical Harmonics (SH) in 3DGS ensures a detailed scene depiction.
However, 3DGS's reliance on explicit geometry increases storage demands, particularly due to the storage of high-order SH parameters for each Gaussian. Moreover, the direct impact of color gradients on Gaussian kernels makes the learning process in 3DGS noise-sensitive and prone to optimization errors. This can lead to an overabundance of redundant Gaussians, increasing storage requirements and negatively affecting training and rendering efficiency, especially in large-scale scenes. Furthermore, the proliferation of high-resolution aerial photography, often exceeding 4K, poses additional challenges for 3DGS in terms of training and rendering of large-scale scenes (see Fig.~\ref{fig:head}).

To tackle the above-outlined issues, this paper introduces an optimized adaptation of 3DGS for efficient large-scale scene representation, termed 'EfficientGS'. This method leverages the representational capacity of individual Gaussians to significantly reduce redundancy in 3DGS, thus alleviating computational and storage demands without sacrificing quality.
'EfficientGS' employs selective densification and pruning of Gaussians. In densification, we focus on non-steady state Gaussians based on their gradient modulus sum, minimizing unnecessary cloning and splitting. This results in fewer Gaussians and enhanced rendering quality. The subsequent pruning phase removes Gaussians with minimal contribution to sight rays, thereby eliminating representational redundancy and optimizing the Spherical Harmonics (SH) of remaining Gaussians.
Furthermore, we introduce a sparse order increment strategy for SH during training. This strategy assesses the need for SH order adjustment based on color disparity from different viewing angles, reducing the SH order where redundant. This not only eases the training burden but also significantly reduces the model size.
Experimental results demonstrate that 'EfficientGS' markedly decreases the Gaussian count, enabling faster training and rendering for high-resolution, large-scale scenes.

In summary, the main contributions of 'EfficientGS' include:
\begin{itemize}
  \item A selective densification strategy that focuses on non-steady state Gaussians, reducing the total number and improving rendering quality.
  \item A pruning strategy that eliminates non-dominant Gaussians, streamlining scene representation by relegating secondary roles to neighboring Gaussians.
  \item A sparse SH order increment strategy during training, reducing computational load and model size by pruning excess SH parameters.
\end{itemize}

\begin{figure*}
    \centering
    \includegraphics[width=1.0\linewidth]{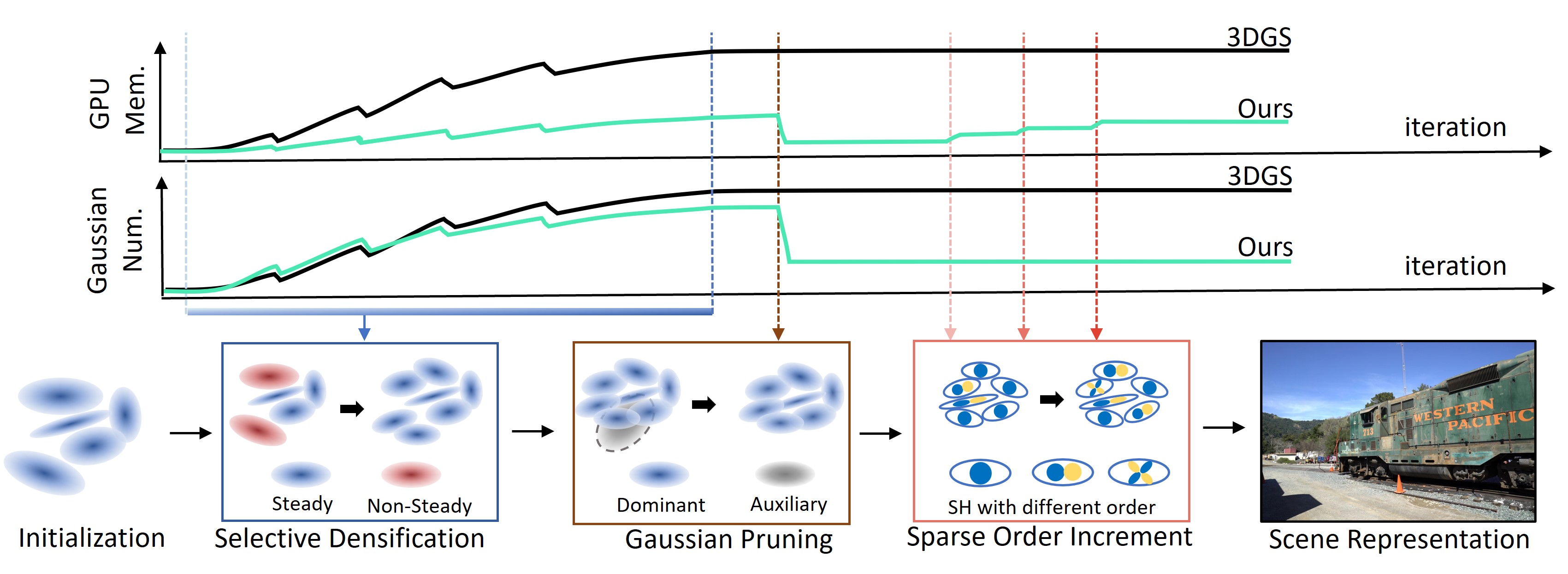}
    \caption{\textbf{Overview of EfficientGS.} As with 3DGS, we use SFM points for initialization, and through differentiable rasterization, we can iteratively optimize the Gaussian parameters using color loss. Every few iterations, we apply a \textcolor[RGB]{68,114,196}{selective densification} strategy(Sec.~\ref{sec3:sgd}), where we determine whether the Gaussian reaches the steady state or not, and the non-steady state Gaussian is densified. After several densifications, we get detailed Gaussians which are fewer in number compared to vanilla 3DGS, but redundancy still exists. So we use the \textcolor[RGB]{139,69,19}{Gaussian pruning} strategy(Sec.~\ref{sec3:gp}) to remove auxiliary Gaussians and continue iterative optimization to hand over the work of these auxiliary Gaussians to the rest Gaussians. During the iteration process, we apply the \textcolor[RGB]{232,114,102}{sparse order increment} strategy(Sec.~\ref{sec3:ssh}) for the SH of Gaussians. We don't start increasing the SH order until after pruning, which greatly reduces the GPU memory pressure. Then, every few iterations, we determine the degree of need for higher-order SH for each Gaussian, depending on which we gradually increase the SH order. Finally, we get our EfficientGS model for scene representation.}
    \label{fig:pipline}
\end{figure*}

\section{Related Works}

\paragraph{Novel view synthesis}
Novel view rendering is a technique that reconstructs a 3D scene using images from different view directions, based on which an image from a novel view can be rendered. The most influential work in this field in recent years is Nerf~\cite{mildenhall2020nerf}, an implicit representation based on neural radiance fields. The success of NeRF can be attributed to its use of positional encoding to enhance the expressive power of the MLP and hierarchical sampling to speed up the rendering process. Much research stems from this vein and has taken neural radiance field rendering to a new level. Mip-NeRF~\cite{barron2021mip} designs an integrated positional encoding (IPE) based on 3D conical frustum defined by a camera pixel to achieve anti-aliased rendering. Mip-NeRF360~\cite{barron2022mip360} further extends Mip-NeRF to better deal with unbounded scenes using a non-linear scene parameterization, online distillation, and a novel distortion-based regularizer. InstantNGP~\cite{InstantNGP} proposes to use multi-resolution hash tables composed of trainable feature vectors that a tiny MLP then processes to achieve substantial training speedup. Zip-NeRF~\cite{barron2023zip} combines the ideas of Mip-NeRF360 and InstantNGP to guarantee high rendering quality and fast training speeds simultaneously. Despite these efforts, NeRF-based methods still have trouble balancing the rendering quality and the computation cost for training and inference. On the other hand, the point-based rendering methods~\cite{zhang2022differentiable,ruckert2022adop} use points to explicitly represent the scene, which has the advantage of drastically reducing the memory consumption as well as increasing the speed of inference and training. Recently, 3DGS~\cite{kerbl20233d} proposes to combine 3D Gaussians with differentiable point-based rendering techniques to realize both fast rendering speed and high quality, which opens up a promising new way for novel view rendering methods. However, also owning to the point-based nature, the storage and computation costs of 3DGS will increase as the number of points increases, leading to a significant decrease in performance and making it hard to deal with large-scale scene representation. Although there exist some recent works~\cite{girish2023eagles,lee2023compact} that attempt to make 3DGS more lightweight, they focus mainly on storage pressure and do not consider much for the training and rendering. 

\paragraph{Large-scale scene representation}
Traditional approaches to obtaining large-scale scene representations consist of several steps: first, the Structure-From-Motion(SFM) technique~\cite{schonberger2016structure} is used to obtain the camera parameters of the captured images, then the Multi-View Stereo technique~\cite{schonberger2016pixelwise,APDMVS,xu2022self} is used to obtain a dense point cloud of the scene, then the surface of the scene is extracted from the point cloud using either implicit~\cite{kazhdan2013screened} or explicit methods~\cite{labatut2009robust} and depending on the quality of the surface mesh, it may be possible to additionally use surface refinement~\cite{li2016efficient} and simplification~\cite{daniels2008quadrilateral}, finally, texture mapping~\cite{waechter2014let} is used to obtain the surface with realistic textures. Such a pipeline can generate a textured mesh to represent the real scene, but it cannot reconstruct non-Lambertian surfaces well because of its extreme dependence on photometric consistency, despite efforts from some works~\cite{jancosek2011multi}, the reconstruction results are still far from satisfactory. To better deal with the complexity of large-scale scenes, such as non-Lambertian surfaces, researchers have tried to apply NeRF~\cite{mildenhall2020nerf}, an implicit scene representation, to the task of large-scale scene representation. Block-nerf~\cite{tancik2022block} decomposes the scene into individually trained NeRFs.This decomposition decouples the rendering time from the scale of the scene, allowing the rendering to scale to arbitrarily large environments and permitting per-block updates. \cite{rematas2022urban} realizes accurate street-level scene representation by incorporating lidar information and sky segmentation with RGB signals. Mega-NeRF~\cite{MegaNerf} first analyzes the visibility statistics for large-scale scenes, based on which a sparsified network structure is designed to improve rendering and training speed. GridNeRF~\cite{xu2023grid} proposed integrating grid-based and NeRF-based approaches to realize a unified scene representation that can solve the blurred renderings on large-scale scenes due to underfitting. Despite the above works having made many improvements to NeRF to fit large-scale scene tasks better, they still can not achieve both high quality and low time consumption for training and rendering.

\section{Method}

\paragraph{Background: 3D Gaussian Splatting}

3DGS is a typical point-based 3D scene representation method and is initialized from sparse point clouds obtained by SFM. Each 3D Gaussian $g$ is defined by its mean, covariance, opacity, and view-dependent color represented by SH. A scene is represented by a set of Gaussians, each defined by a covariance matrix $\Sigma$ centered at point (mean) $\mu$:
\begin{equation}
    G(x) = e^{-\frac{1}{2}(x-\mu)^T\Sigma^{-1}(x-\mu)},
\end{equation}
where $x$ is an arbitrary position in the scene.
To ensure stability in the optimization process and guarantee that the covariance matrix $\Sigma$ is positive semi-definite, it is further decomposed into rotation and scaling components: 
\begin{equation}
    \Sigma = RSS^TR^T.
\end{equation}

During the rendering process, the 3D Gaussians are projected onto the 2D image plane and then combined using alpha blending.
The covariance matrix projection from the 3D space to the 2D image plane is obtained by using the projective transform matrix $P$ (world-to-camera matrix) and the Jacobian $J$ (which represents the approximation of the projective transform). This is described as:
\begin{equation}
    \Sigma' = JP\Sigma P^TJ^T.
\end{equation}

The color $C$ of pixel $p$ is thus computed by blending 3D Gaussian that overlap the given pixel, sorted according to their depth:
\begin{equation}
    C_p=\sum_{i\in \mathcal{N}_p}c_i\alpha_i\prod_{j=1}^{i-1}(1-\alpha_j),
\end{equation}
where $\mathcal{N}_p$ is the ordered set of Gaussians overlapping the pixel $p$, $c$ denotes the view-dependent color computed from SH coefficients.

\begin{figure}
    \centering
    \includegraphics[width=0.85\linewidth]{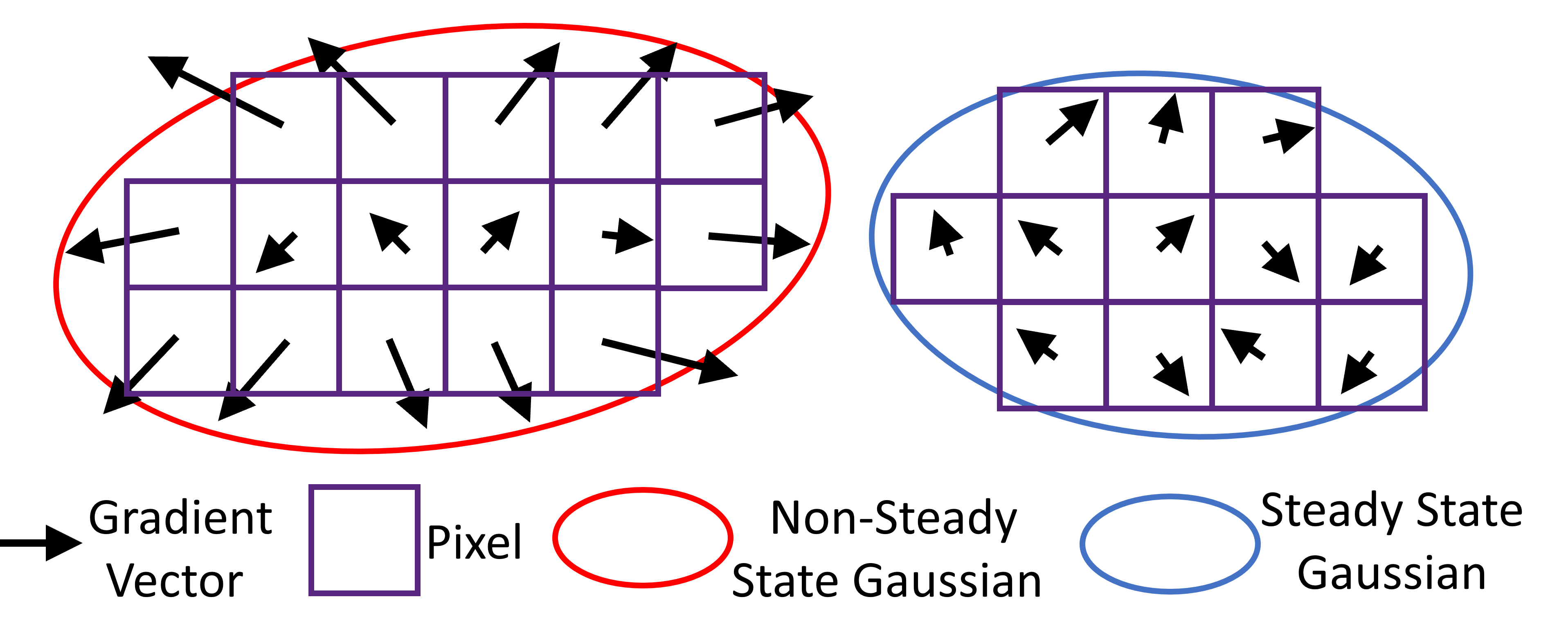}
    \caption{\textbf{Gaussians in different state.} The left \textcolor[RGB]{255,0,0}{red one} is in the non-steady state and the right \textcolor[RGB]{68,114,196}{blue one} is in the steady state. The direction of the straight line arrow represents the direction of the position gradient generated by this pixel for the Gaussian, and the length represents the norm of the gradient vector. For both Gaussians, the view-space position gradients $\sum(\vec\nabla_{p_I}^{g}) \approx 0$, but obviously the red one needs to be densified.}
    \label{fig:efficient_densify}
\end{figure}

\subsection{Selective Gaussian Densification}
\label{sec3:sgd}
The initial Gaussians come from the sparse points of the SFM, which are not sufficient to represent the details of the scene, so the vanilla 3DGS adds an adaptive control strategy to dynamically increase the number of Gaussians. Specifically, if a Gaussian is with an average magnitude $\mathbf{E_g}$ of view-space position gradients above a threshold $\tau_{pos}$, it is considered to be at the regions corresponding to under-reconstruction or over-reconstruction and needs to be densified. 
$\mathbf{E_g}$ is calculated as:
\begin{equation}
  \begin{aligned}
    \mathbf{E}_g =  \frac{\sum_{I\in{M}}^{M}(|\sum(\vec\nabla_{p_{I}}^{g})|)}{|M|},
  \end{aligned}
\end{equation}
where $\vec\nabla_{p_{I}}^{g}$ is the gradient vector for the Gaussian $g$ generated from pixel $p$ on image $I$, $\sum(\vec\nabla_{p_{I}}^{g})$ is the so-called view-space position gradient from image $I$, which is the sum of the gradient vectors of all pixels covered by $g$, and $M$ is a set of images that will produce gradients for $g$. However, such an adaptive control strategy can be problematic. While 3DGS uses a Gaussian to represent multiple pixels on the view space, and if the Gaussian is ideal, the colors of these pixels should be similar to the color of the Gaussian, then $\sum(\vec\nabla_{p_{I}}^{g})\approx0$, which is the basis logic of adaptive control. But pixels and the Gaussian having similar colors is only a sufficient but not necessary condition for $\sum(\vec\nabla_{p_{I}}^{g})\approx0$. As shown in Fig.~\ref{fig:efficient_densify}, the case on the right represents that the Gaussian is already a good representation of the pixels it covers, whereas in the case on the left, it is clear that the Gaussian needs to be densified, but the $\sum(\vec\nabla_{p_{I}}^{g})\approx0$ in both cases. Thus, $\tau_{pos}$ is difficult to determine. A too-high $\tau_{pos}$ can lead to a decrease in the rendering quality since the Gaussian in a case like the one on the left fails to be densified, and a too-low $\tau_{pos}$ would result in a lot of Gaussians in cases like the one on the right are densified, creating a large number of redundant Gaussians and putting a huge strain on both rendering and training. 

To solve this dilemma, we redefine the standard of judgment for Gaussian densification. Specifically, we classify Gaussians into steady state and non-steady state, and a non-steady Gaussian is the one with:
\begin{equation}
  \begin{aligned}
    \mathbf{S}_g  > \tau_{s}, \mathbf{S}_g =  \frac{\sum_{I\in{M}}^{M}(\sum(|\vec\nabla_{p_{I}}^{g}|))}{|M|},
  \end{aligned}
\end{equation}
where $\tau_{s}$ is the threshold to divide the two states. Instead of using the view-space position gradient, we first calculate the norm of the gradient vector from each pixel and then get the sum of them, the size of which can exactly represent whether the Gaussian has reached a converged steady state, i.e., the case of the Gaussian on the right side in Fig.~\ref{fig:efficient_densify}. In this way, we can accurately find these non-steady state Gaussians that need to be densified and achieve more efficient densification to reduce the generation of redundant points while still maintaining the quality of rendering. Then, the same as the vanilla 3DGS, for these Gaussians that need to be densified, we decide whether to clone them or split them based on their scales.

\subsection{Gaussian Pruning}
\label{sec3:gp}


Although we have reduced meaningless densification by using the strategy in Sec.~\ref{sec3:sgd}, during the splitting process in densification, the positions of the newly generated Gaussians are sampled according to the scaled random Gaussian distribution, this still can lead to the generation of many redundant Gaussians. 
Controlling the split positions of these new Gaussians can reduce the redundancy, but it is computationally intensive and limits the degree of freedom of the 3DGS to a certain extent, so instead we propose to add a simpler but effective pruning strategy for deleting the redundant Gaussians after the densification.

Inspired by the NeRF~\cite{mildenhall2020nerf} importance sampling strategy, we can also identify important Gaussians.
Specifically, for a ray emitted from the pixel $p$, we can calculate the weight of a Gaussian $g_i$ when contributing to the pixel $p$ according to:
\begin{equation}
  \begin{aligned}
    weight^i_p=\alpha_i\prod_{j=1}^{i-1}(1-\alpha_j), g_i\in\mathcal{N}_p.
  \end{aligned}
\end{equation}
It is experimentally observed that each ray passes through a large number of Gaussian points, and most of them have very low weights except for some dominant contributors. These dominant Gaussians are of clear geometrical meaning and correspond to real objects in 3D space, whereas those that are not dominant in any ray are only auxiliaries of the adjacent dominant Gaussians. These auxiliaries take on some of the color rendering work that should belong to dominant Gaussians, which corresponds to redundancy in the scene representation. To make scene representation more efficient, we apply our pruning strategy to keep only dominant Gaussians. For each ray, we sort the weights and regard the Gaussians with top-K weights as dominant, that is, we only retain the Gaussian $g$ with:
\begin{equation}
  \begin{aligned}
    weight^g_p \geq sort(\{-weight^j_p, g_j\in\mathcal{N}_p\})[K].
  \end{aligned}
\end{equation}


For all pixels of all images, as long as a Gaussian is dominant for a certain pixel, the Gaussian will be retained. As a result, those auxiliaries will be deleted. After this, we continue to optimize the color of the dominant Gaussians, gradually handing over the work of auxiliaries to the dominant ones.


\begin{table*}
\centering
\resizebox{1.0\linewidth}{!}{
\begin{tabular}{l|cccccc|cccccc|cccccc}
Dataset   & \multicolumn{6}{c|}{Mip-NeRF360}              & \multicolumn{6}{c|}{Tanks\&Temples}           & \multicolumn{6}{c}{Deep Blending}             \\
Method & SSIM$\uparrow$ & PSNR$\uparrow$  & LPIPS$\downarrow$ & Train$\downarrow$  & FPS$\uparrow$  & Storage$\downarrow$   & SSIM$\uparrow$  & PSNR$\uparrow$  & LPIPS$\downarrow$ & Train$\downarrow$  & FPS$\uparrow$  & Storage$\downarrow$   & SSIM$\uparrow$  & PSNR$\uparrow$  & LPIPS$\downarrow$ & Train$\downarrow$  & FPS$\uparrow$  & Storage$\downarrow$   \\ \hline
Plenoxels† & 0.626 & 23.08 & 0.463 & 25m49s & 6.79 & 2.1GB & 0.719 & 21.08 & 0.379 & 25m5s  & 13.0 & 2.3GB & 0.795 & 23.06 & 0.510 & 27m49s & 11.2 & 2.7GB \\
INGP-Base† & 0.671 & 25.30 & 0.371 & \textbf{5m37s}  & 11.7 & 13MB  & 0.723 & 21.72 & 0.330 & \textbf{5m26s}  & 17.1 & 13MB  & 0.797 & 23.62 & 0.423 & \textbf{6m31s}  & 3.26 & 13MB  \\
INGP-Big†  & 0.699 & 25.59 & 0.331 & 7m30s  & 9.43 & 48MB  & 0.745 & 21.92 & 0.305 & 6m59s  & 14.4 & 48MB  & 0.817 & 24.96 & 0.390 & 8m     & 2.79 & 48MB  \\
M-NeRF360† & 0.792 & \textbf{27.69} & 0.237 & 48h    & 0.06 & \textbf{8.6MB} & 0.759 & 22.22 & 0.257 & 48h    & 0.14 & \textbf{8.6MB} & 0.901 & 29.40 & 0.245 & 48h    & 0.09 & \textbf{8.6MB} \\
3DGS†      & 0.815 & 27.21 & \textbf{0.214} & 41m33s & 134  & 734MB & 0.841 & 23.14 & 0.183 & 26m54s & 154  & 411MB & 0.903 & 29.41 & \textbf{0.243} & 36m2s  & 137  & 676MB \\ 
\hline
3DGS*
& 0.812 & 27.48 & 0.222 & 25m57s & 125 & 742MB
& \textbf{0.844} & \textbf{23.64} & \textbf{0.178} & 13m29s      & 164 & 433MB
& 0.900 & 29.54 & 0.247 & 21m56s & 137 & 661MB     \\
C3DGS†     & 0.798     & 27.08     & 0.247     & 33m06s      & 128    & 49MB     & 0.831     & 23.32     & 0.201     & 18m20s      & 185    & 39MB     & 0.901     & 29.79 & 0.258     & 27m33s      & 181    & 43MB     \\
EAGLES†
& 0.810 & 27.15 & 0.240 & 19m59s & 137 & 68MB
& 0.840 & 23.41 & 0.20  & 9m51s  & 244 & 34MB
& \textbf{0.910}  & \textbf{29.91} & 0.250  & 17m29s & 160 & 62MB     \\
Ours &\textbf{0.817}     & 27.38     & 0.216     & 19m41s      & \textbf{218}    & 98MB     & 0.837     & 23.45     & 0.197     & 8m05s      & \textbf{439}   & 33MB     & 0.903     & 29.63     & 0.251     & 14m27s      & \textbf{401}    & 40MB \\
\end{tabular}}
\caption{\textbf{Quantitative results evaluated on \textit{Mip-NeRF 360}, \textit{Tanks\&Temples}, and \textit{Deep Blending} datasets.} Results marked by † represent the results from the papers, and those marked by * represent the results reproduced by ourselves on one A100 GPU using the official code. C3DGS and EAGLES, similarly aiming for lightweight 3DGS, were also tested on one A100. Compared to other methods, our method can simultaneously guarantee high rendering quality, faster training and rendering speed, and lower storage overheads. }
\label{tab:small-scale}
\end{table*}

\subsection{Sparse Order Increment of SH}
\label{sec3:ssh}

After minimizing the number of Gaussians, the computational and storage consumption is still high when dealing with high-resolution large-scale scenes, especially the huge GPU memory consumption may lead to the training not being able to proceed.  This is because the vanilla 3DGS allows each Gaussian to carry a three-order SH to improve the quality of rendering. In fact, it is redundant to use three-order SH for every Gaussian as the ambient light in outdoor scenes is typically of low frequency. To reduce the consumption of redundant SHs, a sparse order increment strategy for SH is applied during the training process. Specifically, we set the SH order to 0 at the beginning, and gradually increase the SH order after our Gaussian pruning, which can significantly reduce the peaks of the GPU memory consumption. The purpose of increasing the order of SH is to represent more complex color models to better show the color differences from different viewpoints. Based on this logic,  we can also calculate the differences between the rendered color with the real color in the whole training views and selectively increase the SH order for only those Gaussians with large differences. The view-based color difference $d_k$ for $k$th Gaussian $g$ is calculated as:
\begin{equation}
  \begin{aligned}
    d_k=\sum_{p=1}^{NHW}\mathbbm{1}(k\in\mathcal{N}_p)weight^k_p|C_{p} - C_{p}^{gt}|,
  \end{aligned}
\end{equation}
where $\mathbbm{1}$ denotes the indicator function, 
$N$ represents the number of training images, $H$,$W$ represent image height, and width, respectively. 
We sort all Gaussians by their differences $d_k$ from largest to smallest and select a certain percentage $r_s$ of them with higher differences. We consider these Gaussians to have a higher demand for higher-order SH and add one order to the SH of these Gaussians. The sparse order increment will be performed every several iterations for a total of 3 times.

\section{Experiments}
\subsection{Implementation Details}
We implemented our method by building on~\cite{kerbl20233d} which
uses a PyTorch framework~\cite{paszke2019pytorch} with a CUDA backend for
the rasterization operation.
All experiments were run on an NVIDIA A100 GPU. For the regular small-scale scenes, we set the number of iteration steps to 30000 and set the iteration step for terminating densification to half of the total iteration steps, the same as the vanilla 3DGS. After 15500 iterations have been performed, we execute the Gaussian pruning and the sparse order increment will be implemented at iteration steps 16000, 17000, and 18000, respectively. For large-scale scenes, which correspond to more images as well as more Gaussian points, we increase the iteration steps to 60,000 to ensure convergence, according to which the number of steps to perform the other operations will be modified proportionately. The $\tau_{s}$ for selective densification is set to $0.0007$, the K for Gaussian pruning is set to 1, and the sparse rate $r_s$ for sparse order increment is set to 0.2 in all experiments.

\begin{figure*}
    \centering
    \includegraphics[width=1\linewidth]{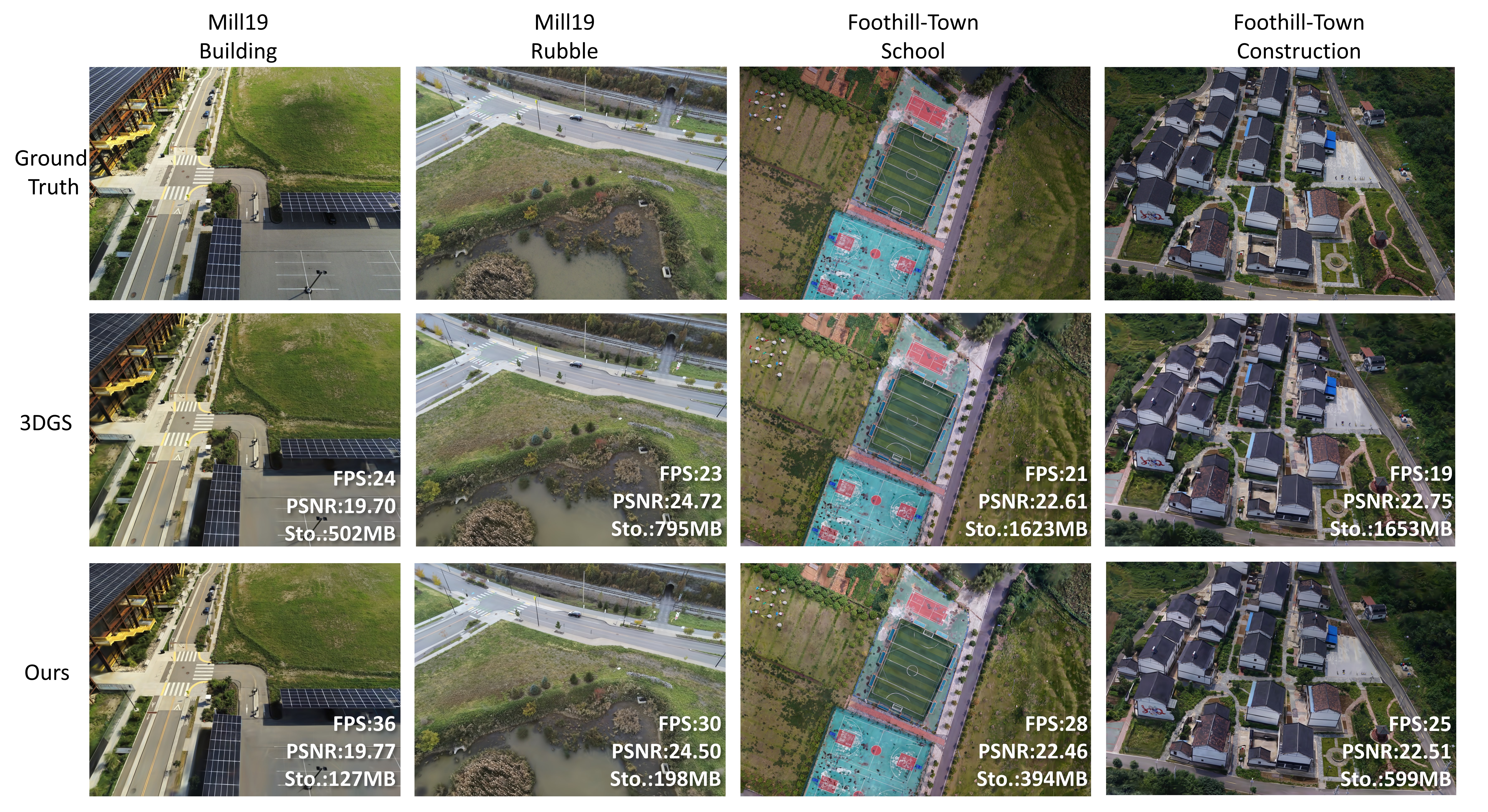}
    \caption{\textbf{Qualitative comparison on 4 scenes of \textit{Mill19} and \textit{Foothill-Town} datasets.} There is nearly no difference between our method and the vanilla 3DGS in terms of rendering quality, but we can achieve higher FPS and lower storage due to the more efficient scene representations our method has learned. Zoom in for more details.}
    \label{fig:compare}
\end{figure*}


\begin{table}
\centering
\resizebox{1.0\linewidth}{!}{
\begin{tabular}{l|cccccc}
Scene& \multicolumn{6}{c}{Building} \\

Method & SSIM & PSNR  & LPIPS & Train  & FPS  & Storage \\
     \hline
Mega-NeRF† & 0.547 & 20.93 & 0.504 & 29h49m & 0.01 & - \\
3DGS* & 0.668 & 19.70 & 0.461 & 7h21m & 24 & 502MB \\ 
Ours & 0.669 & 19.77 & 0.457 & 7h01m & 36 & 127MB \\ 
\hline
Scene& \multicolumn{6}{c}{Rubble}        \\
Method & SSIM & PSNR  & LPIPS & Train  & FPS  & Storage   \\
\hline
Mega-NeRF† & 0.553 & 24.06& 0.516 & 20h48m & 0.01 & - \\
3DGS* & 0.712 & 24.72 & 0.455 & 7h09m & 23 & 795MB \\
Ours & 0.708 & 24.50 & 0.455 & 6h04m & 30 & 198MB \\
\end{tabular}
}
\caption{\textbf{Quantitative results evaluated on \textit{Mill19} dataset.} Note that Mega-NeRF is trained on 8 V100 GPUs, while 3DGS and ours only use one A100 GPU.}
\label{tab:large-scale}
\end{table}

\begin{table}
\centering
\resizebox{1.0\linewidth}{!}{
\begin{tabular}{l|cccccc}
   Scene  & \multicolumn{6}{c}{School}                  \\
   Method  & SSIM & PSNR  & LPIPS & Train  & FPS  & Storage    \\
     \hline
3DGS* & 0.640 & 22.61 & 0.493 & 9h05m & 21 & 1623MB \\
Ours & 0.644 & 22.46 & 0.482 & 8h22m & 28 & 394MB  \\ 
\hline
  Scene   & \multicolumn{6}{c}{Construction}         \\
 Method    & SSIM & PSNR  & LPIPS & Train  & FPS  & Storage   \\
     \hline
3DGS* & 0.606 & 22.75 & 0.498 & 9h34m & 19 & 1653MB \\
Ours & 0.615 & 22.51 & 0.501 & 9h17m & 25 & 599MB \\ 
\end{tabular}}
\caption{\textbf{Quantitative results on the self-collected \textit{Foothill-Town} dataset.}}
\label{tab:large-scale-self}
\end{table}


\begin{table}
\centering
\resizebox{1.0\linewidth}{!}{
\begin{tabular}{ccc|ccccccc}
SD  & GP  & SOI & SSIM & PSNR  & LPIPS & Train  & FPS  & Storage & Peak GPU$\downarrow$ \\ \hline
& & & 0.812 & 27.48 & 0.222 & 25m57s & 125 & 742MB & 12.8G \\
\checkmark & & & 0.820 & 27.53 & 0.202 & 24m02s & 140 & 544MB & 11.0GB \\
\checkmark & \checkmark & & 0.821 & 27.48 & 0.209 & 20m42s  & 229 & 259MB & 10.9GB\\
\checkmark & \checkmark & \checkmark & 0.817 & 27.38 & 0.216     & 19m41s & 218 & 98MB & 8.8GB
\end{tabular}}
\caption{\textbf{Ablation study on Mip-NeRF360.} SD represents selective densification, GP represents Gaussian pruning, SOI represents sparse order increment for SH.}
\label{tab:ablation}
\end{table}

\begin{table}
\centering
\resizebox{1.0\linewidth}{!}{
\begin{tabular}{ccccccc}
 \multicolumn{1}{c|}{TopK} & SSIM & PSNR  & LPIPS & Train  & FPS  & Storage  \\ \hline
\multicolumn{1}{c|}{1}    & 0.817&	27.38&	0.216&	19m41s&	218& 98MB\\
\multicolumn{1}{c|}{2}    & 0.818 & 27.43& 0.210&20m44s & 185 & 123MB \\
\multicolumn{1}{c|}{3}    & 0.818 & 27.45& 0.209&21m07s & 174 & 137MB \\
\end{tabular}}
\caption{\textbf{Ablation study for Gaussian pruning on Mip-NeRF360.} We use different K values for the TopK in Gaussian pruning in our full method to test the impact of K.}
\label{tab:ablation_k}
\end{table}

\begin{table}
\centering
\resizebox{1.0\linewidth}{!}{
\begin{tabular}{ccccccccc}
 SR & Random &\multicolumn{1}{c|}{Ours} & SSIM  & PSNR  & LPIPS & Train  & FPS  & Storage   \\ \hline
0.1  & \checkmark &\multicolumn{1}{c|}{}    & 0.812 & 27.02& 0.220&19m38s & 211 & 71MB \\
0.1  & &\multicolumn{1}{c|}{\checkmark}       & 0.815&	27.27&	0.218&	19m43s&	215&	78MB \\  \hline
0.2  & \checkmark  &\multicolumn{1}{c|}{}    & 0.814 & 27.12& 0.218&19m56s & 211 & 84MB  \\
0.2  & &\multicolumn{1}{c|}{\checkmark}       & 0.817&	27.38&	0.216&	19m41s&	218&	98MB  \\  \hline
0.5  & \checkmark  &\multicolumn{1}{c|}{}    & 0.818 & 27.28& 0.214&20m13s & 206 & 134MB \\ 
0.5  & &\multicolumn{1}{c|}{\checkmark}       & 0.819&	27.46&	0.213&	19m36s&	220&	158MB \\

\end{tabular}}
\caption{\textbf{Ablation study for sparse order increment of SH on Mip-NeRF360.} SR representes Sparse Rate $r_s$.}
\label{tab:ablation_sr}
\end{table}

\subsection{Datasets and Metrics}

We validate the effectiveness as well as the generalization ability of our method on multiple datasets, including common small-scale scenes and large-scale scenes that we focus on.
First, in order to demonstrate that our method is generalizable and works well with common small-scale datasets, we test our method on the same datasets as the vanilla 3DGS, including 9 scenes from the \textit{Mip-Nerf360} dataset~\cite{barron2022mip}, 2 scenes from the \textit{Tanks\&Temples} dataset~\cite{knapitsch2017tanks}, and 2 scenes from the \textit{Deep Blending} dataset~\cite{hedman2018deep}. 
Similarly, we use every 8th view for evaluation, while the remaining views are utilized for training. 

Then coming back to our focus, we test our approach on 2 scenes (namely the ~\textit{Building} and the ~\textit{Rubble}) from \textit{Mill19} dataset~\cite{MegaNerf}, a public high-resolution large-scale dataset with a resolution of $4608 \times 3456$.


In addition, we also collect a town-scale dataset, termed \textit{Foothill-Town}, with a space size of $2km \times 2km$. This dataset contains 15,000 images with a resolution of $6000 \times 4000$, which are captured by 5 cameras on a drone following a grid flight path.

The primary metrics used to assess reconstruction quality are PSNR (Peak Signal-to-Noise Ratio), SSIM (Structural Similarity Index), and LPIPS (Learned Perceptual Image Patch Similarity)~\cite{zhang2018unreasonable}.
In addition, we also evaluated the model's rendering speed, training time, peak GPU memory consumption during training, and storage cost, which are also important for large-scale scene representation.
\subsection{Experiments Results}

\textbf{Common datasets.} We first compare our method with other SOTA methods (Plenoxel~\cite{fridovich2022plenoxels}, INGP~\cite{InstantNGP}, M-NeRF360~\cite{barron2022mip360}, 3DGS~\cite{kerbl20233d}, C3DGS~\cite{lee2023compact}, EAGLES~\cite{girish2023eagles}) on common small-scale datasets. As can be seen from Tab.~\ref{tab:small-scale}, compared with the vanilla 3DGS, our method can achieve faster rendering, training speed, and reduced storage consumption to the order of magnitude of the NeRF-based method, while guaranteeing the same level of rendering quality, resulting in a more efficient scene representation. And compared to methods for 3DGS compacting~\cite{lee2023compact}, they consume slightly less storage than we do, but at the cost of longer training times, and more degradation of rendering quality. Also, since they don't reduce much of the number of Gaussians, they don't get enough increase in rendering speed. Thus, they are not competent for the task of large-scale scene representation.

\noindent\textbf{Large-scale datasets.} 
 We compare our method with the vanilla 3DGS, as well as MegaNeRF~\cite{MegaNerf}, a SOTA NeRF-based method for large-scale scenes. We use Colmap~\cite{schonberger2016structure} to get the SFM results required for vanilla 3DGS and our method. In particular, since the time consumption of Colmap goes up massively when too many images are involved, we sampled the images of ~\textit{Building} evenly, so we only used 1/2 of the images for training. Nonetheless, For ~\textit{Building}, our method still achieves the same level of rendering quality compared to MegaNeRF, which uses the full image for training. For ~\textit{Rubble}, we use full images for training and can achieve better quality than MegaNeRF. Meanwhile, MegaNeRF performs poorly in terms of training speed and rendering speed on either scene under such rendering quality.
 Compared with the vanilla 3DGS, with our more efficient scene representation, we can achieve real-time rendering and faster training speed, which is important for large-scale scene representation. 
 As shown in Fig.~\ref{fig:compare}, in terms of the rendered image, our rendering quality is almost the same as the original Gaussian.

To further demonstrate our good performance on high-resolution large-scale datasets, we also compare our method with the vanilla 3DGS on our \textit{Foothill-Town} dataset. Specifically, we extract two scenes (\textit{School} and \textit{Construction}, each contains over 1000 images) from this dataset and feed them to Colmap to obtain SFM results. As with the vanilla Gaussian, we take one-eighth of the images as the validation set, and Tab.~\ref{tab:large-scale-self} shows the comparison of our method with the vanilla 3DGS on the validation set of these two scenes. Our approach ensures real-time rendering even when processing 6K images.


\subsection{Ablation Studies}
\textbf{Selective Gaussian densification.}
As can be seen from Tab.~\ref{tab:ablation}, using our selective densification strategy can not only reduce the number of Gaussians but also improve the rendering quality. This is because our strategy prevents the decrease in quality due to the splitting of the steady-state Gaussians. At the same time, due to the reduction of the number of Gaussians, the computational pressure will be alleviated accordingly, resulting in faster training and rendering as well as lower peak GPU memory overhead, which makes our method more friendly to large-scale high-resolution scene representation. 
In addition, fewer Gaussians also correspond to less storage overhead.

\noindent\textbf{Gaussian pruning.}
Our Gaussian pruning strategy can dramatically reduce the number of Gaussians, while the rendering quality nearly remains the same. This leads to a significant improvement in rendering speed, allowing our approach to guarantee real-time rendering for large-scale scene representation. We also conduct a detailed experiment for the choice of K for retaining dominant Gaussians. As shown in Tab.~\ref{tab:ablation_k}, a smaller K represents more stringent filtering, and therefore more Gaussians will be removed, but also brings a certain degree of degradation in rendering quality. Conversely, when K is increased, more Gaussians are retained and the rendering quality will increase accordingly. Weighing rendering quality and speed, we chose to set K to 1 in the final version for better handling of large-scale scenes.


\noindent\textbf{Sparse order increment for SHs.}
We start with the order of SH equal to 0 and only increase the SH order after Gaussian pruning, which can reduce the GPU memory overhead as shown in Tab.~\ref{tab:ablation}, making it possible to train scenes with a larger scale under the GPU memory limitation. Finally, our SH sparsification can significantly reduce storage overhead. We also compare the effect of different $r_s$ on the results, as well as comparing our method of selecting the Gaussian that needs to increase the SH order with random selection in Tab.~\ref{tab:ablation_sr}. Compared to random selection, our method guarantees better rendering quality. Increasing $r_s$ improves the quality a bit, and when $r_s=0.5$, the quality is basically the same as before, which proves that there is indeed a lot of redundancy in SH. To better reduce the storage pressure, we set $r_s$ to 0.2 in our experiments.



\section{Conclusion}
In this paper, we propose EfficientGS, an efficient scene representation method based on 3D Gaussian Splatting, which can better handle high-resolution large-scale scenes. 
We first design a selective densification strategy, which selectively densifies non-steady state Gaussians with a large magnitude of conflicts inside. 
This targeted densification significantly reduces redundant processing, preserves converged Gaussians, and ultimately enhances rendering quality while reducing Gaussian count.
We then Gaussian pruning, where we keep only the dominant Gaussians on each ray to absorb the work of those auxiliary Gaussians with no real geometric meaning. 
Finally, We implement sparse order increment for SH to selectively assign higher-order SH to Gaussians based on their color deviation from ground truth across views.
As a result, our approach simultaneously guarantees fast training, high-speed rendering, low storage consumption, low GPU memory overhead, and high rendering quality, as well as great generalization validated on a variety of datasets, including lots of large-scale scenes.

\textbf{Limitation and future work.} The same as the vanilla 3DGS, the time to execute our different modules is all related to the iteration steps only, while the optimal execution timing should be different when training scenes of different scales since different numbers of images are involved, the wrong module execution time will lead to the degradation of the overall performance. In our future work, we will design strategies to compute the optimal execution timing of different modules to improve the robustness of our EfficientGS.


\appendix


\bibliographystyle{named}
\bibliography{ijcai24}

\end{document}